\RequirePackage{etoolbox}
\csdef{input@path}{%
 {}% cls, sty files
 {img/}% eps files
}%
\csgdef{bibdir}{}% bst, bib files

\documentclass[ba]{imsart}
\pubyear{2016}
\volume{00}
\issue{0}
\doi{0000}
\firstpage{1}
\lastpage{1}

\usepackage{natbib}

\DeclareRobustCommand{\df}{\mathrm{d}}

\newcommand{\g}{\,\vert\,}
\newcommand{\E}{\mathbb{E}}
\newcommand{\var}{\mathrm{Var}}

\newcommand{\mult}{\mathrm{Mult}}
\newcommand{\dir}{\mathrm{Dir}}

\newcommand{\expect}[2]{\mathbb{E}_{#1}\left[ #2 \right]}

\newcommand{\mcL}{\mathcal{L}}
\newcommand{\mcN}{\mathcal{N}}
\newcommand{\lda}{\mathrm{LDA}}

\newcommand{\bx}{\mathbf{x}}
\newcommand{\bz}{\mathbf{z}}
\newcommand{\bw}{\mathbf{w}}
\newcommand{\xnew}{x_{\mathrm{new}}}

\newcommand{\mbbeta}{\mb{\beta}}

\newcommand{\braces}[1]{\left\{ #1 \right\}}
\newcommand{\parenths}[1]{\left( #1 \right)}
\newcommand{\sq}[1]{\left[ #1 \right]}

\newcommand{\expfam}{\textsc{ExpFam}}
\newcommand{\ovexpfam}{\textsc{OvExpFam}}

\usepackage[algoruled]{algorithm2e}

\usepackage[T1]{fontenc}
\usepackage{amssymb, amsfonts}

\DeclareRobustCommand{\mb}[1]{\boldsymbol{#1}}

\usepackage{amsmath}
\usepackage{bm}

\usepackage[english]{babel}
\usepackage[parfill]{parskip}  % ?
\usepackage{afterpage}         % ?
\DeclareRobustCommand{\parhead}[1]{\textbf{#1}~}

\usepackage{lineno}
\usepackage{ragged2e}

\usepackage[usenames,dvipsnames]{xcolor}
\definecolor{shadecolor}{gray}{0.9}
\definecolor{shadecolor}{gray}{0.9}

\usepackage{mdwlist}

\usepackage{enumitem}

\usepackage{graphicx}
\usepackage[labelfont=bf]{caption}
\usepackage[format=hang]{subcaption}

\usepackage{natbib}
\usepackage{hyperref}
\usepackage{cleveref} % must come after hyperref
\creflabelformat{equation}{#1#2#3}

\usepackage{framed}

\usepackage{amsthm}
\usepackage{amsmath}

\startlocaldefs
\numberwithin{equation}{section}
\theoremstyle{plain}

\endlocaldefs

\begin{document}

\begin{frontmatter}
\title{A General Method for \\ Robust Bayesian Modeling}
\runtitle{A General Method for Robust Bayesian Modeling}
% \thankstext{T1}{Footnote to the title with the ``thankstext'' command.}

\begin{aug}
\author{\fnms{Chong} \snm{Wang}\thanksref{addr1}\ead[label=e1]{chowang@microsoft.com}}
and
\author{\fnms{David M.}
\snm{Blei}\thanksref{addr2}\ead[label=e2]{david.blei@columbia.edu}}

\address[addr1]{Microsoft Research, One Microsoft Way, Redmond, WA
  98052, \printead{e1} % print email address of "e1"
}

\address[addr2]{Department of Statistics and Department of Computer
  Science, Columbia University, New York, NY 10025, \\ \printead{e2}
}

%\thankstext{t1}{Some comment}
%\thankstext{t2}{First supporter of the project}
%\thankstext{t3}{Second supporter of the project}

\end{aug}

\begin{abstract}%   <- trailing '%' for backward compatibility of .sty file

  Robust Bayesian models are appealing alternatives to standard models, providing protection from data that contains outliers or other departures from the model assumptions. Historically, robust models were mostly developed on a case-by-case basis; examples include robust linear regression, robust mixture models, and bursty topic models. In this paper we develop a general approach to robust Bayesian modeling. We show how to turn an existing Bayesian model into a robust model, and then develop a generic strategy for computing with it.  We use our method to study robust variants of several models, including linear regression, Poisson regression, logistic regression, and probabilistic topic models. We discuss the connections between our methods and existing approaches, especially empirical Bayes and James-Stein estimation.

\end{abstract}


\end{frontmatter}
\section{Introduction}

% XXX should this introduction have a more concrete example? it's
% pretty abstract. [dmb]

% XXX note that here we go from robustness to the predictive
% distribution; in the main text i think we go from the predictive
% distribution to robustness. we should think about what is more
% natural. [dmb]

% TODO [after submission] there seems to be a debate---discussed in
% the encyclopedia of statistics---about whether "models" or "methods"
% are robust. here we are declaring models to be robust, giving a
% general method for creating robust models, and giving a unified
% method for fitting them.  another point: a benefit of bayesian
% modeling is generic algorithms for analyzing data---we now get the
% benefits of this for robust modeling.  let's talk about this
% somewhere, maybe in the introduction. [dmb]

Modern Bayesian modeling enables us to develop custom methods to
analyze complex data~\citep{Gelman:2003,Bishop:2006,Murphy:2013}. We
use a model to encode the types of patterns we want to discover in the
data---either to predict about future data or explore existing
data---and then use a posterior inference algorithm to uncover the
realization of those patterns that underlie the observations.  Modern
Bayesian modeling has had an impact on many fields, including natural
language processing~\citep{Blei:2003b,Teh:2006c}, computer
vision~\citep{Fei-Fei:2005}, the natural
sciences~\citep{Pritchard:2000}, and the social
sciences~\citep{Grimmer:2009}.  Innovations in scalable inference
allow us to use Bayesian models to analyze massive
data~\citep{Hoffman:2013,Welling:2011,Ahn:2012,Xing:2013}; innovations
in generic inference allow us to easily explore a wide variety of
models~\citep{Ranganath:2014,Wood:2014,Hoffman:2014}.

But, as George Box famously quipped, all models are
wrong~\citep{Box:1976}.  Every Bayesian model will fall short of
capturing at least some of the nuances of the true distribution of the
data.  This is the important problem of model mismatch, and it is
prevalent in nearly every application of modern Bayesian modeling.
(Even if a model is not wrong in theory, which is rare, it is often
wrong in practice, where some data are inevitably corrupted such as by
measurement error or other problems.)

One way to cope with model mismatch is to refine our model, diagnosing
how it falls short and trying to fix its issues~\citep{Gelman:1996}.
But refining the model ad infinitum is not a solution to model
mismatch---taking the process of model refinement to its logical
conclusion leaves us with a model as complex as the data we are trying
to simplify. Rather, we seek models simple enough to understand the
data and to generalize from it, but flexible enough to accommodate its
natural complexity.  These are models that both discover important
predictive patterns in the data and flexibly ignore unimportant
issues, such as outliers due to measurement error. Of course, this is
a trade off: A model that is too flexible will fail to generalize; a
model that is too rigid will be lead astray by unsystematic deviations
in the data.

To develop appropriately flexible procedures, statisticians have
traditionally appealed to \textit{robustness}, the idea that
inferences about the data should be ``insensitive to small deviations
from the assumptions.''~\citep{Huber:2009}.  The goal of robust
statistics is to safeguard against the kinds of deviations that are
too difficult or not important enough to diagnose. One popular
approach to robust modeling is to use M-estimators~\citep{Huber:1964},
where the basic idea is to reweigh samples to account for data
irregularity.  Another approach, which we build on here, is to replace
common distributions with heavy-tailed distributions, distributions
that allow for extra dispersion in the data.  For example, this is the
motivation for replacing a Gaussian with a student's t in robust
linear regression~\citep{Lange:1989,Fernandez:1999,Gelman:2003} and
robust mixture modeling~\citep{Peel:2000,Svensen:2005}.  In discrete
data, robustness arises via contagious distributions, such as the
Dirichlet-multinomial, where seeing one type of observation increases
the likelihood of seeing it again. For example, this is the type of
robustness that is captured by the bursty topic model
of~\citet{Doyle:2009}.

Robust models are powerful, but each must be developed on a
case-by-case basis.  Beginning with an existing non-robust model, each
requires a researcher to derive a specific algorithm for a robust
version of that model.  This is in contrast to more general Bayesian
modeling, which has evolved to a point where researchers can often
posit a model and then easily derive a Gibbs sampler or variational
inference algorithm for that model; robust modeling does not yet enjoy
the same ease of use for the modern applied researcher.

Here we bridge this gap, bringing the idea of robust modeling into
general Bayesian modeling.  We outline a method for building a robust
version of any Bayesian model and derive a generic algorithm for
computing with them.  We demonstrate that our methods allow us to
easily build and use robust Bayesian models, models that are less
sensitive to inevitable deviations from their underlying assumptions.

\parhead{Technical summary.}  We use two ideas to build robust
Bayesian models: localization and empirical Bayes.  At its core, a
Bayesian model involves a parameter $\beta$, a likelihood
$p(x_i \g \beta)$, a prior over the parameter $p(\beta \g \alpha)$,
and a hyperparameter $\alpha$.  In a classical Bayesian model all data
are assumed drawn from the parameter (which is assumed drawn from the
prior),
\begin{align}
  \label{eq:classical-bayes}
  p(\beta, \bx \g \alpha) = p(\beta \g \alpha) \prod_{i=1}^{n} p(x_i \g \beta).
\end{align}
To make it robust, we turn this classical model into a localized
model.  In a localized model, each data point is assumed drawn from an
individual realization of the parameter $p(x_i \g \beta_i)$ and that
realization is drawn from the prior $p(\beta_i \g \alpha)$,
\begin{align}
  \label{eq:localized-model}
  p(\mbbeta, \bx \g \alpha) = \prod_{i=1}^{n} p(\beta_i \g \alpha) p(x_i \g
  \beta_i).
\end{align}
This is a more heterogeneous and robust model because it can explain
unlikely data points by deviations in their individualized parameters.
Of course, this perspective is not new---it describes and generalizes
many classical distributions that are used for robust modeling.  For
example, the student's t distribution, Dirichlet-multinomial
distributions, and negative Binomial distribution all arise from
marginalizing out the local parameter $\beta_i$ under various prior
distributions.

But there is an issue.  The model of \Cref{eq:classical-bayes} uses
the parameter $\beta$ to share information across data points; the
hyperparameter can be safely fixed, and often is in many applications
of Bayesian models.  In the localized model of
\Cref{eq:localized-model}, however, each data point is independent of
the others.  To effectively share information across data we must fit
(or infer) the hyperparameter $\alpha$.

Fitting the hyperparameter is a type of empirical Bayes
estimation~\citep{Robbins:1964,Copas:1969,Efron:1973,Efron:1975a,Robbins:1980,Maritz:1989,Carlin:2000,Carlin:2000a}.
The general idea behind empirical Bayes is to use data to estimate a
hyperparameter.  Specifically, setting $\alpha$ to maximize
\Cref{eq:localized-model} gives a parametric empirical Bayes estimate
of the prior on
$\beta_i$~\citep{Morris:1983,Kass:1989}.\footnote{Alternatively, we
  can put priors on $\alpha$ to seek maximum a prior (MAP) estimate or
  full Bayesian treatment.  In this paper however, for simplicity, we
  focus on the maximum likelihood estimate.  We emphasize that all
  these choices retain a model of uncertainty around the parameter
  $\beta$.}  While localization is not typically part of the empirical
Bayes recipe---one can just as easily fit the hyperparameters to
maximize the marginal probability of the data under the original
Bayesian model---hierarchical models of the form of
\Cref{eq:localized-model} appear extensively in the empirical Bayes
literature~\citep{Efron:1973,Efron:1975a,Morris:1983,Kass:1989,Efron:1996,Carlin:2000a,Efron:2010a}.
Other names for the localized model include a compound sampling model,
a two-stage sampling model, and an empirical Bayes model.  For
perspective on empirical Bayes, see the excellent review
by~\cite{Carlin:2000}.

We develop a general algorithm to optimize the hyperparameter $\alpha$
to maximize the likelihood of the data in the robust model.  This
algorithm is a central contribution of this paper; it generalizes the
case-by-case algorithms of many existing robust models and expands the
idea of robustness to a wider class of models, including those that
rely on approximate posterior inference.  As a demonstration, we use
our strategy to study robust generalized linear
models~\citep{McCullagh:1989}---linear regression, Poisson regression,
and logistic regression---as well as robust topic
models~\citep{Blei:2003b,Doyle:2009}.  We find that robust Bayesian
models enjoy improved predictive performance and better estimates of
unknown quantities.

\parhead{Organization of this paper.} \Cref{sec:method} briefly
reviews classic Bayesian modeling and introduces the idea of
localization to robustify a Bayesian model. \Cref{sec:models} presents
several examples of how to apply this idea, developing robust variants
of exponential family models, generalized linear models, and topic
models. \Cref{sec:algorithms} describes how to compute with a robust
Bayesian model using expectation maximization and nonconjugate
variational inference. Finally, \Cref{sec:study} reports results with
several models on both synthetic and real data.

%% FIXME "latent variables" and "parameters" are inconsistent. [cw]

%%% Local Variables:
%%% mode: latex
%%% TeX-master: "ba-submit"
%%% End:

\section{A General Method for Robust Bayesian Modeling}
\label{sec:method}

\begin{figure}[t]
  \begin{center}
    (a)
    \includegraphics[width=0.25\textwidth]{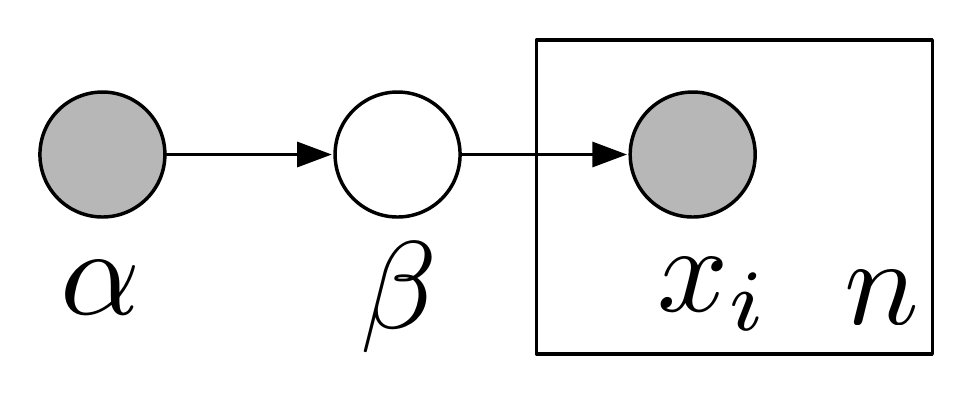}
    (b)
    \includegraphics[width=0.25\textwidth]{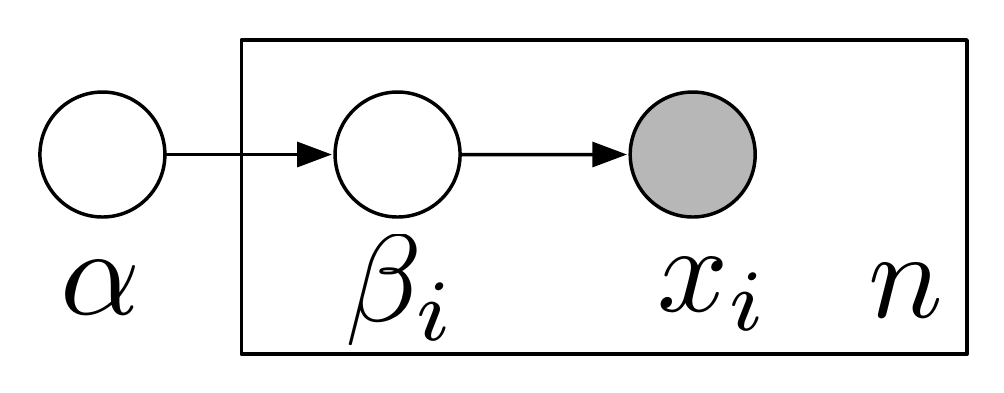}
  \end{center}

  \caption{(a) A graphical model for a standard Bayesian model (b) A
    graphical model for a localized model. By drawing a new parameter
    $\beta_i$ for each data point, the localized model allows
    individual data points to vary more than the standard Bayesian
    model as in (a).  Note the robust model \textit{requires} that we
    fit the hyperparameter $\alpha$.  Otherwise the data are rendered
    completely independent.}
  \label{fig:gm}
\end{figure}

We first describe standard Bayesian modeling, and the key ingredients
that we will build on.  We then develop robust Bayesian modeling.

\subsection{Bayesian models}

Bayesian modeling uses probability to capture uncertainty around
unknown parameters in a statistical model~\citep{Gelman:2003}. A
Bayesian model is a joint distribution of parameters $\beta$ and a
data set $\bx$. In an exchangeable model, this joint factorizes into a
product of likelihood terms for each data point $p(x_i \g \beta)$ and
a prior of the parameters $p(\beta \g \alpha)$
(\Cref{eq:classical-bayes}).  The prior is governed by the
hyperparameter $\alpha$.

\Cref{fig:gm} (a) shows the graphical model.  This model easily
generalizes to include conditional models, such as in Bayesian linear
regression and logistic regression~\citep{Bishop:2006}, local latent
variables, such as in Bayesian mixtures~\citep{Adrian:2001} and topic
models~\citep{Blei:2003b,Blei:2012}, and non-exchangeable data, such
as in a hidden Markov model~\citep{Rabiner:1989} or Kalman
filter~\citep{Kalman:1960}.  For now we focus on the simplest setting
in~\Cref{eq:classical-bayes}.

When we use a model we condition on data and compute the corresponding
posterior, the conditional distribution of the parameter given the
data. We then employ the posterior in an application, such as to form
predictions or to investigate properties of the data.

The posterior is proportional to the product of the prior and the data
likelihood,
\begin{align}
  p(\beta \g \bx, \alpha) \propto p(\beta \g \alpha) \prod_{i=1}^{n} p(x_i
  \g \beta).
  \label{eq:posterior}
\end{align}
We can use the posterior to form the posterior predictive
distribution, the distribution of a new data point conditional on the
observed data.  The posterior predictive distribution is
\begin{align}
  \label{eq:posterior-predictive}
  p(\xnew \g \bx, \alpha) = \int p(\xnew \g \beta) p(\beta \g \bx,
  \alpha) \df \beta.
\end{align}
It integrates the data likelihood $p(\xnew \g \beta)$ under the
posterior distribution $p(\beta \g \bx, \alpha)$.  The posterior
predictive distribution is an important idea in Bayesian modeling.  It
is used both to form predictions about the future and to check,
diagnose, and select
models~\citep{Geisser:1979,Rubin:1984,Gelman:1996}.

% When the number of data is small then the prior plays an important
% role in the posterior; with more data the likelihood dominates the
% prior. As the number of data gets large, the posterior becomes peaked
% and converges to a point mass~\citep{Gelman:2003}.  (Technically, this
% is only true in finite-dimensional models.)  Note this property is
% agnostic to the ``truth''---the posterior will become peaked even if
% the data did not truly come from the model.

\subsection{Robust Bayesian models}
\label{sec:robust-models}

One of the virtues of Bayesian modeling is that the model's prediction
about the future does not rely on a single point estimate of the
parameter, and averaging over the posterior can mitigate overfitting.
However, the Bayesian pipeline does not explicitly aim for a good
predictive distribution on future data---the posterior predictive of
\Cref{eq:posterior-predictive} is the true distribution of unseen data
only when the chosen model represents the true distribution of the
data~\citep{Bernardo:1994}.

In practice, we use models to simplify a complex data generating
process~\citep{Box:1980}.  Thus there is always a mismatch between the
posterior predictive distribution and the true distribution of future
data.  This motivates the philosophy behind robust
statistics~\citep{Huber:2009}, which aims to develop methods that are
not sensitive to small changes in the modeling assumptions.

We develop robust Bayesian models, models that can usefully
accommodate deviation from the underlying assumptions. We will use two
ideas: localization and empirical Bayes.

\parhead{The localized model.}  The first idea is
\textit{localization}.  As we discussed above, a traditional Bayesian
model independently draws the data $\bx$ conditional on the parameter
$\beta$, which is drawn from the prior.  The localized model relaxes
this to a hierarchical model, which governs each data point with an
individual parameter $\beta_i$ that is drawn from the prior.
In~\Cref{eq:localized-model}, the joint distribution of the
individualized parameters $\mbbeta = \beta_{1:n}$. The localized model
captures heterogeneity in the data; it explains atypical data that
deviates from the norm (i.e., outliers) by deviations in their
parameters.

One way to view the localized model is as one where each data point is
drawn independently and identically distributed (IID) from an
integrated likelihood,
\begin{align}
  p(x_i \g \alpha) = \int p(x_i \g \beta_i) p(\beta_i \g \alpha)
  \df \beta_i.
\end{align}
When the parameter $\beta_i$ controls the dispersion of the data, this
results in a heavier-tailed distribution than the original
(unintegrated) observation model in~\Cref{eq:classical-bayes}.  

For example, suppose the original model assumes data are from a
Gaussian distribution.  Localizing the variance under an inverse gamma
prior reveals the student's t-distribution, a commonly-used
distribution for adding robustness to models of continuous
data~\citep{Gelman:2003}.  In \Cref{sec:models} we show that many
methods in robust statistics can be interpreted in this way.

% XXX challenge: can we prove that with the exponential family and
% conjugate prior, the integrated likelihood has a heavy tail, or at
% least, prove general conditions on this?  (i mention this again in
% the robust modeling section.) [dmb]

% [cw]: still couldn't figure it out a good way to show this. however,
% i changed some of the sentences to soften this claim.

% XXX i added some intuitive language, around the dispersion being
% localized.

\parhead{Empirical Bayes estimation.} Crucially, localization requires
that we infer the hyperparameter---it is now $\alpha$ that carries
common information about the data.  We can see this
graphically. Localization takes us from the traditional Bayesian model
in \Cref{fig:gm} (a) to the model in \Cref{fig:gm} (b).  In
\Cref{fig:gm} (b), fixing $\alpha$ renders the data completely
independent.

We fit the hyperparameter with maximum likelihood.  This amounts to
finding a prior on $\beta_i$ that maximizes the integrated likelihood
of the data in in the robust model,
\begin{align}
  \label{eq:integrated-likelihood}
  \hat{\alpha} = \arg \max_{\alpha} \sum_{i=1}^{n} \log \int p(x_i \g \beta_i) p(\beta_i \g
  \alpha) \df \beta_i.
\end{align}
Here we marginalize out the individualized parameters $\beta_i$ from
\Cref{eq:localized-model}.  Thus fitting a robust model implicitly
optimizes the predictive distribution. 

%One recently proposed related work, population empirical
%Bayes~\citep{Kucukelbir:2015}, also optimizes the predictive
%distribution to mitigate model misspecification.)  

Directly optimizing the likelihood can be difficult because of the
integral inside the log function.  We defer this issue to
\Cref{sec:algorithms}, where we show how to optimize with a
combination of variational inference~\citep{Jordan:1999} and the
expectation maximization algorithm~\citep{Dempster:1977}.  Our
approach---localizing a global variable and then fitting its
prior---allows us to develop robust variants of many Bayesian models.
As we mentioned in the introduction, this perspective has close
connections to empirical
Bayes~\citep{Efron:1973,Efron:1975a,Efron:2010a} and the empirical
Bayes approach laid out in~\cite{Carlin:2000a}.

% XXX commented out

% Maximizing \Cref{eq:integrated-likelihood} is the main computational
% problem for robust probabilistic models, and the main contribution
% is a generic method for solving it. Our method involves the
% expectation-maximization algorithm~\citep{Dempster:1977} and
% variational inference~\citep{Jordan:1999,Wainwright:2008},
% especially a recent extension to nonconjugate
% models~\citep{Wang:2012}. This principle---localizing a global
% variable and then using our method to fit its prior---

% XXX repetitive; commented out.

% Note that fitting the hyperparameter is a form of empirical
% Bayes---indeed, localized models themselves are often referred to as
% empirical Bayes models.  The perspective of this paper has close
% connections to the empirical Bayes perspective on James-Stein
% estimation~\citep{Efron:1973,Efron:1975a,Efron:2010a} and the
% empirical Bayes approach laid out in~\cite{Carlin:2000a}.

% XXX commented out; we don't need this. [dmb]

% Finally, we note that empirical Bayes ideas have been criticized for
% using the data to both fit the local parameters $\beta_i$ and the
% prior $\alpha$.  In fact, this ``double use'' of the data occurs
% irrespective of the estimation procedure.  Even a fully Bayesian
% analysis, where we place a fixed prior on $\alpha$, implicitly uses
% the data for both purposes.  See the excellent review
% by~\cite{Carlin:2000} for details.}

\parhead{The predictive distribution.}  With a localized model and
estimated hyperparameter, we form predictions about future data with
the corresponding predictive distribution
\begin{align*}
  p(x^* \g \hat{\alpha}) = \int p(x^*\g \beta^*) p(\beta^* \g \hat{ \alpha})
  {\rm d} \beta^*.
\end{align*}
Notice this has the same form as the likelihood term in the objective
of \Cref{eq:integrated-likelihood}.

This predictive procedure motivates our approach to localize
parameters and then fit hyperparameters with empirical Bayes.  One
goal of Bayesian modeling is to make better predictions about unseen
data by using the integrated likelihood, and the traditional Bayesian
approach of \Cref{eq:posterior-predictive} is to integrate relative to
the posterior distribution. For making predictions, however, the
traditional Bayesian approach is mismatched because the posterior is
not formally optimized to give good predictive distributions of each
data point.  As we mentioned, it is only the right procedure when the
data comes from the model~\citep{Bernardo:1994}.

In contrast, the robust modeling objective of
\Cref{eq:integrated-likelihood}---the objective that arises from
localization and empirical Bayes---explicitly values a distribution of
$\beta$ that gives good predictive distributions for each data point,
even in the face of model mismatch.

%In this section we presented a general approach to building robust
%Bayesian models.  We next outline several examples
%(\Cref{sec:models}), a generic algorithm for computing with them
%(\Cref{sec:algorithms}), and a case study that demonstrates the
%improved performance of robust models (\Cref{sec:study}).
%
% XXX note: kass:1989 takes laplace approximations of the
% hyperparameter, i.e., taylor expansions around the MAP. (we don't
% necessarily need to say this in our text.  i'm just noting it here.)
% [dmb]

\section{Practicing Robust Bayesian Modeling}
\label{sec:models}

Machine learning and Bayesian statistics have produced a rich
constellation of Bayesian models and general algorithms for
computing about them~\citep{Gelman:2003,Bishop:2006,Murphy:2013}. We
have described an approach to robustifying Bayesian models in
general, without specifying a model in particular.  The recipe is to
form a model, localize its parameters, and then fit the
hyperparameters with empirical Bayes.  In \Cref{sec:algorithms}, we
will develop general algorithms for implementing this procedure.
First we describe some of the types of models that an investigator may
want to make robust, and give some concrete examples.

First, many models contain hidden variables within $p(x_i \g \beta)$,
termed \textit{local variables} in \cite{Hoffman:2013}.  Examples of
local variables include document-level variables in topic
models~\citep{Blei:2003b}, component assignments in mixture
models~\citep{McLachlan:2000}, and per-data point component weights in
latent feature models~\citep{Salakhutdinov:2008a}.  We will show how
to derive and compute with robust versions of Bayesian models with
local hidden variables.  For example, the bursty topic models
of~\cite{Doyle:2009} and the robust mixture models of~\cite{Peel:2000}
can be seen as variants of robust Bayesian models.

Second, some models contain two kinds of parameters, and the
investigator may only want to localize one of them.  For example the
Gaussian is parameterized by a mean and variance. Robust Gaussian
model need only localize the variance; this results in the student's
t-distribution. In general, these settings are straightforward.
Divide the parameter into two parts $\beta = [\beta_1, \beta_2]$ and
form a prior that divides similarly $\alpha = [\alpha_1, \alpha_2]$.
Localize one of the parameters and estimate its corresponding
hyperparameter.

% XXX commented out this footnote.  it confused the story more than
% added to it. [dmb]

% \footnote{In fact, localizing the mean under a Gaussian prior will
%   only lead to an increase in variance and will not be a robust
%   model. However, with fixed variance in the original Gaussian model,
%   it leads to a similar approach as the James-Stein estimation, which
%   indeeds results in better predictions. We will detail this in
%   Section~\ref{sec:models}.}

Last, many models are not fully generative, but draw each data point
conditional on covariates.  Examples include linear regression,
logistic regression, and all other generalized linear
models~\citep{McCullagh:1989}.  This setting is also straightforward
in our framework. We will show how to build robust generalized linear
models, such as Poisson regression and logistic regression, and how to
fit them with our algorithm.

% XXX i am here

We now show how to build robust versions of several types of
Bayesian models.  These models connect to existing robust methods
in the research literature, each one originally developed on a
case-by-case basis. 
%In contrast, our approach and algorithms generalize to many models.

\subsection{Conjugate exponential families}
\label{sec:conjugate}

% TODO [after submission] an idea to fold into this subsection is
% shrinkage.  the posterior expectation of each data point's theta
% depends on the hyperparameter.  when we fit the hyperparameter then
% each data point inference depends on all the other data. [dmb]

The simplest Bayesian model draws data from an exponential family and
draws its parameter from the corresponding conjugate prior.  The
density of the exponential family is
\begin{align*}
  \label{eq:expfam}
  p(x \g \eta) = h(x) \exp \braces{\eta^\top t(x) - a_x(\eta)},
\end{align*}
where $t(x)$ is the vector of sufficient statistics, $\eta$ is the
natural parameter, and $h(x)$ is the base measure.  The log normalizer
$a_x(\eta)$ ensures that the density integrates to one,
\begin{align}
  a_x(\eta) = \int \exp\braces{\eta^\top t(x)} \df x.
\end{align}
The density is defined by the sufficient statistics and natural
parameter.  When $x_i$ comes from an exponential family we use the
notation $x_i \sim \expfam(\eta, t(x))$.

Every exponential family has a conjugate prior~\citep{Diaconis:1979}.
Suppose the data come from $x_i \sim \expfam(\eta, x)$, i.e., the
exponential family where $x$ is its own sufficient statistic.  The
conjugate prior on $\eta$ is
\begin{align*}
  p(\eta \g \alpha) = h(\eta) \exp\braces{\alpha^\top [\eta, -a_x(\eta)]
    - a_\eta(\alpha)}.
\end{align*}
This is an exponential family whose sufficient statistics concatenate
the parameter $\eta$ and the negative log normalizer $-a_x(\eta)$ in
the likelihood of the data.  The parameter divides into two components
$\alpha = [\alpha_1, \alpha_2]$ where $\alpha_1$ has the same
dimension as $\eta$ and $\alpha_2$ is a scalar.  Note the difference
between the two log normalizers: $a_x(\eta)$ normalizes the data
likelihood; $a_\eta(\alpha)$ normalizes the prior.  In our notation,
$\eta \sim \expfam([\alpha_1, \alpha_2], [\eta, -a_x(\eta)])$.

Given data $\bx$, the posterior distribution of $\eta$ is in the same
exponential family as the prior,
\begin{align}
  \eta \g \bx, \alpha \sim \expfam\parenths{[\alpha_1 + \textstyle
  \sum_i x_i, \alpha_2 + n], [\eta, -a_x(\eta)]}.
\end{align}
This describes the general set-up behind all commonly used conjugate
prior-likelihood pairs, such as the Beta-Bernoulli,
Dirichlet-Multinomial, Gamma-Poisson, and others.  Each of these
models first draws a parameter $\eta$ from a conjugate prior, and then
draws $n$ data points $x_i$ from the corresponding exponential family.

Following \Cref{sec:robust-models}, we define a generic localized
conjugate exponential family,
\begin{align*}
  \eta_i &\sim \expfam\parenths{[\alpha_1, \alpha_2], [\eta,
           -a_x(\eta)]} \\
  x_i &\sim \expfam\parenths{\eta_i, x}.
\end{align*}
We fit the hyperparameters $\alpha$ to maximize the likelihood of the
data in \Cref{eq:integrated-likelihood}.  In a conjugate
exponential-family pair, the integrated likelihood has a closed form
expression.  It is a ratio of normalizers,
\begin{align}
p(x_i \g \alpha) &= \int p(x_i \g \eta) p(\eta \g \alpha) \df \eta
    \nonumber \\
                 &= \exp\braces{a_{\eta}([\alpha_1 + x_i, \alpha_2 +
                   1]) - a_{\eta}([\alpha_1, \alpha_2])}.
                   \label{eq:exp-fam-predictive}
\end{align}

In this setting the log likelihood of \Cref{eq:integrated-likelihood}
is
\begin{align*}
  \mathcal{L}(\alpha_1, \alpha_2; \bx) =
  \left(\textstyle \sum_{i=1}^{n} a_{\eta}(\alpha_1 + x_i, \alpha_2 + 1)\right) - n
  a_{\eta}(\alpha_1, \alpha_2).
\end{align*}

This general story connects to specific models in the research
literature. As we described above, it leads to the student's
t-distribution when the data come from a Gaussian with a fixed mean
and localized variance $x_i \sim \mcN(\mu, \sigma_i^2)$, and when the
variance $\sigma^2_i$ has an inverse Gamma prior.  It is ``robust''
when the dispersion parameter is individualized; the model can explain
outlier data by a large dispersion.  Fitting the hyperparameters
amounts to maximum likelihood estimation of the student's
t-distribution.

% TODO [after submission]: consider another example, e.g.,
% negative-binomial from the poisson? [dmb]

This simple exchangeable model also connects to James-Stein
estimation, a powerful method from frequentist statistics that can be
understood as an empirical Bayes
procedure~\citep{Efron:1973,Efron:2010a}.  Here the data are from a
Gaussian with fixed variance and localized mean
$x_i \sim \mcN(\mu_i, \sigma^2)$, and the mean $\mu_i$ has a Gaussian
prior $\mu_i \sim \mcN(0, \lambda^2)$.  (This is the conjugate prior.)
We recover a shrinkage estimate similar to James-Stein estimation by
fitting the prior variance with maximum likelihood.

\subsection{Generalized linear models}
\label{sec:glm}

Generalized linear models (GLM) are conditional models of a response
variable $y$ given a set of covariates $x$~\citep{McCullagh:1989}.
Specifically, canonical GLMs assume the response is drawn from an
exponential family with natural parameters equal to a linear
combination of coefficients $w$ and covariates.
\begin{align*}
  \eta_i &= w^\top x_i \\
  y_i &\sim \expfam(\eta_i, y).
\end{align*}
Many conditional models are generalized linear models; some of the
more common examples are linear regression, logistic regression, and
Poisson regression.  For example, Poisson regression sets
$\eta_i = w^\top x_i$ to be the log of the rate of the Poisson
distribution of the response.  (This fits our notation---the log of
the rate is the natural parameter of the Poisson.)

We use the method from \Cref{sec:robust-models} to construct a robust
GLM.  We replace the deterministic natural parameter with a Gaussian
random variable,
\begin{align*}
  \eta_i &\sim \mcN(w^\top x_i, \lambda^2).
\end{align*}
Its mean is the linear combination of coefficients and covariates, and
we fit the coefficients $w$ and variance $\lambda^2$ with maximum
likelihood. This model captures heterogeneity among the response
variables.  It accommodates outliers and enables more robust
estimation of the coefficients.

Unlike \Cref{sec:conjugate}, however, this likelihood-prior pair is
typically not conjugate---the conditional distribution of $\eta_i$
will not be a Gaussian and the integrated likelihood is not available
in closed form.  We will handle this nonconjugacy with the algorithms
in \Cref{sec:algorithms}.

In our examples we will always use Gaussian priors. However, we can
replace them with other distributions of the reals.  We can interpret
the choice of prior as a regularizer on a per-data point ``shift
parameter.''  This is the idea behind \cite{She:2011} (for linear
regression) and \cite{Tibshirani:2013} (for logistic regression).
These papers set up a shift parameter with $L_1$ regularization,
which corresponds to a Laplace prior in the models described here.

% XXX below, commented out from the last draft. if included we need
% more details (but no math) about what the connection is.  (i need to
% look up this paper.) [dmb]

% This method has strong connections to generalized linear mixed
% models~\citep{Breslow:1993}. Say XXX. Next we provide two concrete
% examples of generalized linear models.

We give two examples of robust generalized linear models: robust
logistic regression and robust Poisson regression.  (We discuss robust
linear regression below, when we develop robust overdispersed GLMs.)

\parhead{Example: Robust logistic regression.} In logistic regression
$y_i$ is a binary response,
\begin{align*}
  y_i \sim \mathrm{Bernoulli}(\sigma( w^\top x_i))
\end{align*}
where $\sigma(t) = (1+\exp(-t))^{-1}$ is the logistic function; it
maps the reals to the unit interval.

We apply localization to form robust logistic regression.  The model is
\begin{align*}
  \eta_i &\sim  \mcN(w^\top x_i, \lambda^2) \\
  y_i &\sim  \mathrm{Bernoulli}(\sigma(\eta_i)),
\end{align*}
where we estimate $w$ and $\lambda^2$ by maximum likelihood.  This
model is robust to outliers in the sense that the per-data
distribution on $\eta_i$ allows individual data to be
``misclassified'' by the model. As we mentioned for the general case,
the Gaussian prior is not conjugate to the logistic likelihood; we can
use the approximation algorithm in~\Cref{sec:algorithms} to compute
with this model.

We note that there are several existing variants of robust logistic
regression.  \citet{pregibon:1982} and \citet{Stefanski:1986} use
M-estimators~\citep{Huber:1964} to form more robust loss functions,
which are designed to reduce the contribution from possible outliers.
Our approach can be viewed as a likelihood-based robust loss function,
where we integrate the likelihood over the individual parameter
$\beta_i$.  This induces uncertainty around individual observations,
but without explicitly defining the form of a robust loss.

Closer to our method is the shift model of~\citet{Tibshirani:2013},
who use $L_1$ regularization, as well as the more recent theoretical
work of~\citet{Feng:2014}.  However, none of this work estimates
hyperparameters $\lambda$.  Using empirical Bayes to estimating such
hyperparameters is at the core of our procedure, and we found in
practice that it is an important component.

\parhead{Example: Robust Poisson regression.}  The Poisson
distribution is an exponential family on positive integers.  Its
parameter is a single a positive value, the rate.  Poisson
regression is a conditional model of a count-valued response,
\begin{align*}
  y_i \sim \mathrm{Poisson}\parenths{\exp\braces{w^\top x_i}}.
\end{align*}

Using localization, a robust Poisson regression model is
\begin{align}
  \eta_i &\sim  \mcN(w^\top x_i, \lambda^2)
  \label{eq:robust-poisson-1} \\
  y_i &\sim \mathrm{Poisson}\parenths{\exp\braces{\eta_i}}. \label{eq:robust-poisson-2}
\end{align}
As for all the models above, this allows individual data points to
deviate from their expected value.  Notice this is particularly
important when the data are Poisson, where the variance equals the
mean.  In classical Poisson regression, the mean is
$\E[Y_i \g x_i] = w^\top x_i$ and the variance is
$\var[Y_i \g x_i] = w^\top x_i$. Here we can marginalize out the
per-data point parameter in the robust model to reveal a larger
variance,
\begin{align*}
  \E[Y_i \g x_i] &= \exp\braces{w^\top x_i + \lambda^2/2} \\
  \var[Y_i \g x_i] &= \exp\braces{w^\top x_i + \lambda^2/2} +
                     \left(\exp\braces{\lambda^2} - 1\right) \exp\braces{2 w^\top x_i + \lambda^2}.
\end{align*}
These marginal moments comes from the fact that
$\exp\braces{w^\top x_i}$ follows a log normal.  Intuitively, as the
prior variance $\lambda^2$ goes to zero they approach the moments for
classical Poisson regression.

% TODO: chong can you check the paragraphs below?  i expanded on the
% discussion of negative binomial regression [dmb]

% FIXTODO: i'm not quite sure what the last sentence in this paragraph
% means here. two reasons: 1) looks like the fitting glm.nb doesn't
% fit the entire gamma distribution parameters 2) the data is
% generated from log normal not from gamma distribution. perhaps i
% should generate the noise from gamma distribution and
% re-compare. actually i will do it.

% XXX yes, what i was trying to say is that our fitting the
% hyperparameter leads to better performance.  but i agree that we
% also changed the distribution from which the localized parameter is
% drawn.  drawing from a gamma is a good idea. [but no need to hold up
% submission on this experiment...] -- dmb

Robust Poisson regression relates to negative binomial
regression~\citep{Cameron:2013}, which also introduces per-data
flexibility. Negative binomial regression is
\begin{align*}
  \epsilon_i &\sim \mathrm{Gamma}(a, b) \\
  \eta_i &= w^\top x_i + \log \epsilon_i \\
  y_i &\sim \textrm{Poisson}(\exp\braces{\eta_i}).
\end{align*}
In this notation, this model assumes that $\epsilon_i$ drawn from a
Gamma distribution and further estimates its parameters with empirical
Bayes. In our study, we found that the robust Poisson model defined
in~\Cref{eq:robust-poisson-1,eq:robust-poisson-2} outperformed
negative binomial regression; we suspect this is because of the
empirical Bayes step. (See \Cref{sec:study}.)

\subsection{Overdispersed generalized linear models}
\label{sec:overdispersed}

An overdispersed exponential family extends the exponential family
with a dispersion parameter, a positive scalar that controls the
variance.  An overdispersed exponential family is
\begin{align}
  p(y \g \eta, \tau) = h(y, \tau)\exp\braces{\frac{\eta^\top t(y) -
  a_y(\eta)}{\tau}},
\end{align}
where $\tau$ is the dispersion.  We denote this
$y \sim \ovexpfam(\eta, t(y), \tau)$.  One example of an overdispersed
exponential family is the Gaussian---the parameter $\eta$ is the mean
and $\tau$ is the variance.  (We can also form a Gaussian in a
standard exponential family form, where the natural parameter combines
the mean and the variance.)

An overdispersed GLM draws the response from an overdispersed
exponential family~\citep{Jorgensen:1987}.  Following
Section~\ref{sec:conjugate}, we localize the dispersion parameter
$\tau$ to create a robust overdispersed GLM.  In this case we draw
$\tau$ from a Gamma,
\begin{align*}
  \tau_i &\sim \textrm{Gamma}(a, b) \\
  y_i &\sim \ovexpfam(w^\top x_i, y_i, \tau_i)
\end{align*}
Localizing the dispersion connects closely with our intuitions around
robustness. An outlier is one that is overdispersed relative to what
the model expects; thus a per-data point dispersion parameter can
easily accommodate outliers.

For example consider the GLM that uses a unit-variance Gaussian with
unknown mean (an exponential family).  This is classical linear
regression.  Now form the overdispersed GLM---this is linear
regression with unknown variance---and localize the dispersion
parameter under the Gamma prior.  Marginalizing out the per-data
dispersion, this model draws the response from a student's t,
\begin{align*}
  y_i \sim  t_{2a}(\cdot \g w^\top x_i, 1/(ab))
\end{align*}
where the student's t notation ${\rm t}_{\nu}(y \g \mu, \phi)$ is
\begin{align*}
  t_{\nu}(y|\mu, \phi) = \frac{\Gamma(\frac{\nu +
  1}{2})}{\Gamma(\frac{\nu}{2})\sqrt{\pi\nu\phi}}
  \left(1+\frac{1}{\nu}\frac{(y-\mu)^2}{\phi}\right)^{-\frac{\nu+1}{2}} .
\end{align*}
This is a robust linear model, an alternative parameterization of the
model of~\citet{Lange:1989} and \citet{Fernandez:1999}.

Intuitively, localized overdispersed models lead to heavy-tailed
distributions because it is the dispersion that varies from data point
to data point.  When working with the usual exponential family (as in
\Cref{sec:conjugate} and \Cref{sec:glm}), the heavy-tailed distribution arises
only when the dispersion is contained in the natural parameter; note
this is the case for our previous examples, logistic regression and
Poisson regression.  Here, the dispersion is localized by design.

% [cw] in addition, section 3.2 and 3.3 did two different ways of
% adding uncertainty to w^x, one is random location and the other is
% random scale / maybe we should explicitly say this a bit?

% XXX good idea.  i added some language about this; see above.  [dmb]

\subsection{Generative models with local and global variables}
\label{sec:complex-models}

We have described how to build robust versions of simple
models---conjugate prior-exponential families, generalized linear
models, and overdispersed generalized linear models.  Modern machine
learning and Bayesian statistics, however, has developed much more
complex models, using exponential families and GLMs as components in
structured joint distributions~\citep{Bishop:2006,Murphy:2013}.
Examples include models of time series, hierarchies, and
mixed-membership. We now describe how to use the method of
Section~\ref{sec:conjugate} build robust versions of such models.

Each complex Bayesian model is a joint distribution of hidden and
observed variables.  \cite{Hoffman:2013} divide the variables into two
types: local variables $\bz$ and global variables $\beta$. Each
local variable $z_i$ helps govern of its associated data point $x_i$
and is conditionally independent of the other local variables. In
contrast, the global variables $\beta$ help govern the distribution of all the
data.  This is expressed in the following joint,
\begin{align*}
  p(\beta, \bz, \bx) = p(\beta) \prod_{i=1}^{n} p(z_i, x_i \g \beta).
\end{align*}
This joint describes a wide class of models, including Bayesian
mixture models~\citep{Ghahramani:2000, Attias:2000}, hierarchical
mixed membership
models~\citep{Blei:2003b,Erosheva:2007,Airoldi:2007a}, and Bayesian
nonparametric models~\citep{Antoniak:1974,Teh:2006b}.\footnote{Again
  we restrict ourselves again to exchangeable models.
  Non-exchangeable models only contain global variables, such as time
series models~\citep{Rabiner:1989,Fine:1998,Fox:2011b, Paisley:2009}
and models for network analysis~\citep{Airoldi:2007,Airoldi:2009}.}

% TODO: above, pepper this with citations. [dmb]
% [cw]: done but have a q about time series model.
% [dmb] what is your question?
% FIXTODO: don't know how to exactly extend to time-series models.
% OK, i've removed the comment that "these models can also be made robust."

\begin{figure}[t]
\begin{center}
  \centerline{\includegraphics[width=0.4\textwidth]{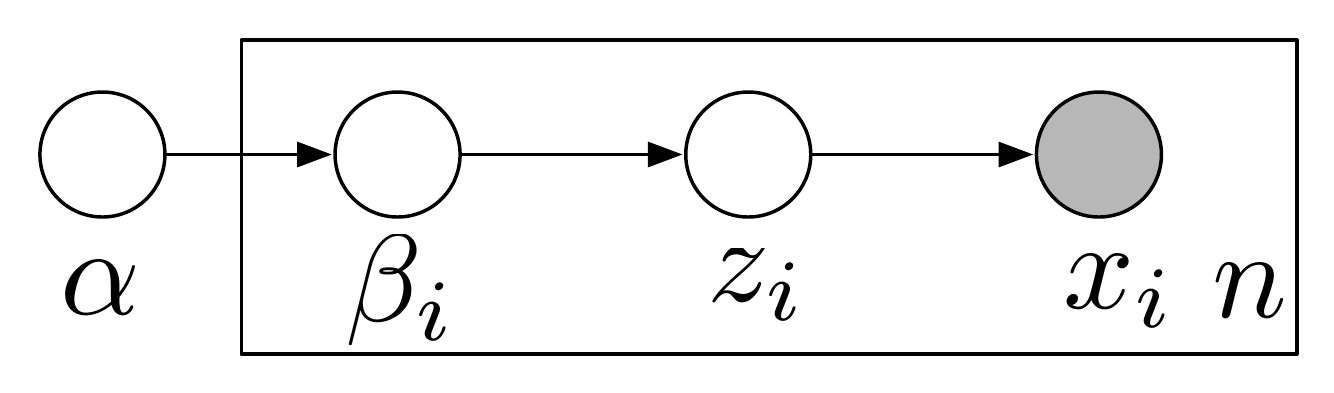}}
  \caption{Robust modeling with with local latent variable $\beta_i$,
    $z_i$ and observation $x_i$, $i=1,\dots,n$.}
  \label{fig:robust-latent-gm}
\end{center}
\end{figure}

To make a robust Bayesian model, we localize some of its global
variables---we bring them inside the likelihood of each data point,
endowing each with a prior, and then fit that prior with empirical
Bayes.  Localizing global variables accommodates outliers by allowing
how each data point expresses the global patterns to deviate from the
norm.  \Cref{fig:robust-latent-gm} shows the graphical model where
$\beta_i$ is the localized global variable and $z_i$ is the original
local variable.

As an example, consider latent Dirichlet allocation
(LDA)~\citep{Blei:2003b}.  LDA is a mixed-membership model of a
collection of documents; each document is a collection of words.  LDA
draws each document from a mixture model, where the mixture
proportions are document-specific and the mixture components (or
``topics'') are shared across the collection.

Formally, define each topic $\beta_k$ to be a distribution over a
fixed vocabulary and fix the number of topics $K$.  LDA assumes that a
collection of documents comes from the following process:
\begin{enumerate}[noitemsep]
\item Draw topic $\beta_k \sim \dir(\eta)$ for $k=1,2,\cdots, K$.
\item For each document $d$,
  \begin{enumerate}[noitemsep]
  \item Draw topic proportions $\theta_d \sim \dir(\alpha)$.
  \item For each word $n$,
    \begin{enumerate}[noitemsep]
    \item Draw topic assignment $z_{dn} \sim \mult(\theta_d)$.
    \item Draw word $w_{dn} \sim \mult(\beta_{z_{dn}})$.
    \end{enumerate}
  \end{enumerate}
\end{enumerate}
The local variables are the topic assignments and topic proportions;
they are local to each document.  The global variables are the topics;
they are involved in the distribution of every document.

For simplicity, denote steps (a) and (b) above as
$\bw_{d} \sim \lda(\mbbeta, \alpha)$, where
$\mbbeta = \{\beta_1, \ldots, \beta_K\}$.  To make LDA robust, we
localize the topics.  Robust LDA still draws each document from a
mixture of $K$ topics, but the topics are themselves drawn anew for
each document.  

Each per-document topic $\beta_{dk}$ is drawn from its own
distribution with a ``master'' topic parameter $\eta_k$, which
parameterizes the Dirichlet of the $k$-th topic.  Localized LDA draws
each document from the following process:
\begin{enumerate}[noitemsep]
\item Draw per-document topic $\beta_{dk} \sim \dir(\eta_k)$, for
  $k=1,2,\cdots, K$.
\item Draw $\bw_d \sim \lda(\mbbeta_d, \alpha)$.
\end{enumerate}
We fit the hyperparameters $\eta_k$, the corpus-wide topics. In the
generative process they are perturbed to form the per-document topics.

This robust LDA model is equivalent to the topic model proposed in
\cite{Doyle:2009}, which accounts for ``burstiness'' in the
distribution of words of each documents.  Burstiness, also called
contagion, is the idea that when we see one word in a document we are
more likely to see that word again.  It is a property of the marginal
distribution of words when integrating out a Dirichlet distributed
multinomial parameter. This is called a Dirichlet-multinomial compound
distribution~\citep{Madsen:2005}.

Burstiness is a good property in topic models.  In a traditional topic
model, repeated terms provide increased evidence for the importance of
that term in its topic.  In contrast, the bursty topic model can
partly explain repeated terms by burstiness.
Consequently, the model does not overestimate that term's importance
in its topic.

LDA is just one example.  With this method we can build robust
versions for mixtures, time-series models, Bayesian nonparametric
models, and many others. As for GLMs, we have a choice of what to
localize.  In topic models we localized the topics, resulting in a
bursty topic model.  In other cases we localize dispersion parameters,
such as in robust Gaussian mixtures~\citep{Svensen:2005}.

\section{Fitting robust Bayesian models}
\label{sec:algorithms}

We have shown how to robustify a wide class of Bayesian models.  The
remaining question is how to analyze data with them.  We now show how
to adapt existing approximate inference algorithms to compute with
robust Bayesian models.  We provide a general strategy that can
be used with simple models (e.g., conjugate exponential families),
nonconjugate models (e.g., generalized linear models), and complex
models with local and global variables (e.g., LDA).

The key algorithmic problem is to fit the hyperparameter in
\Cref{eq:integrated-likelihood}.  We use an expectation-maximization
(EM) algorithm to fit the model~\citep{Dempster:1977}.  In many cases,
some of the necessary quantities are intractable to compute.  We
approximate them with variational
methods~\citep{Jordan:1999,Wainwright:2008}.

Consider a generic robust Bayesian model. The data come from an
exponential family and the parameter from a general prior,
\begin{align*}
  \beta_i &\sim p(\cdot \g \alpha) \\
  x_i &\sim \expfam(x_i, \beta_i).
\end{align*}
Note this is not necessarily the conjugate prior.  Following
\Cref{sec:robust-models}, we fit the hyperparameters $\alpha$
according to \Cref{eq:integrated-likelihood} to maximize the marginal
likelihood of the data.

We use a generalization of the EM algorithm, derived via variational
methods.  Consider an arbitrary distribution of the localized
variables $q(\beta_{1:n})$.  With Jensen's inequality, we use this
distribution to bound the marginal likelihood.  Accounting for the
generic exponential family, the bound is
\begin{align}
  \mathcal{L}(\alpha) & \geq  
  \sum_{i=1}^{n} \E_q\sq{\log p(x_i \g \beta_i)p(\beta_i \g
  \alpha)}-\E_q\sq{\log q(\beta_i)} \nonumber \\ & = \sum_{i=1}^{n}
    \E_q\sq{\beta_i}^\top x_i - \E_q\sq{a(\beta_i)} + \E_q\sq{\log
    p(\beta_i \g \alpha)} - \E_q\sq{\log q(\beta_i)}.
    \label{eq:em-objective}
\end{align}
This is a variational bound on the marginal
likelihood~\citep{Jordan:1999}, also called the ELBO (``the Evidence
Lower BOund'').  Variational EM optimizes the ELBO by coordinate
ascent---it iterates between optimizing with respect to
$q(\beta_{1:n})$ and with respect to the hyperparameters $\alpha$.

Optimizing \Cref{eq:em-objective} with respect to $q(\beta_{1:n})$
minimizes the Kullback-Leibler divergence between $q(\beta_{1:n})$ and
the exact posterior $p(\beta_{1:n} \g \bx)$.

In a localized model, the posterior factorizes,
\begin{align*}
  p(\beta_{1:n} \g \bx, \alpha) = \prod_{i=1}^{n} p(\beta_i \g x_i,
  \alpha).
\end{align*}
Each factor is a posterior distribution of the per-data point
parameter, conditional on the data point and the hyperparameters.  If
each posterior factor is computable then we can perform an exact
E-step, where we set $q(\beta_i)$ equal to the exact posterior.  In
the context of empirical Bayes models, this is the algorithm suggested
by~\cite{Carlin:2000a}.

In many cases the exact posterior will not be available. In these
cases we use variational inference~\citep{Jordan:1999}.  We set
$q(\beta_i)$ to be a parameterized family of distributions over the
$i$th variable $\beta_i$ and then optimize \Cref{eq:em-objective} with
respect to $q(\cdot)$.  This is equivalent to finding the
distributions $q(\beta_i)$ that are closest in KL divergence to the
exact posteriors $p(\beta_i \g x_i, \alpha)$.  It is called a
variational E-step.

The M-step maximizes \Cref{eq:em-objective} with respect to the
hyperparameter $\alpha$.  It solves the following optimization
problem,
\begin{align}
  \hat{\alpha} = \arg \max_{\alpha} \sum_{i=1}^{n} \E_q\sq{\log
  p(\beta_i \g \alpha)}.
\end{align}
At first this objective might look strange---the data do not appear.
But the expectation is taken with respect to the (approximate)
posterior $p(\beta_i \g x_i, \alpha)$ for each localized parameter
$\beta_i$; this posterior summarizes the $i$th data point.  We solve
this optimization with gradient methods.

\parhead{Nonconjugate models.}  As we described, the E-step
amounts to computing $p(\beta_i \g x_i, \alpha)$ for each data point.
When the prior and likelihood form a conjugate-pair
(\Cref{sec:conjugate}) then we can compute an exact
E-step.\footnote{In this setting we can also forgo the EM algorithm
  and directly optimize the marginal likelihood with gradient
  methods---the integrated likelihood is computable in
  conjugate-exponential family pairs (\Cref{eq:exp-fam-predictive}).}  For many
models, however, the E-step is not computable and we need to
approximate $p(\beta_i \g x_i, \alpha)$.

One type of complexity comes from nonconjugacy, where the prior is not
conjugate to the likelihood.  As a running example, robust GLM models
(\Cref{sec:glm}) are generally nonconjugate. (Robust linear regression
is an exception.)  In a robust GLM, the goal is to find optimal
coefficients $w$ and variance $\lambda^2$ that maximizes the robust
GLM ELBO,
\begin{align}
  \label{eq:GLM-objective}
  \mcL(w, \alpha) =
  \sum_{i=1}^{n} \E_q\sq{\eta_i}^\top y_i - \E_q\sq{a(\eta_i)} + \E_q\sq{\log
  p(\eta_i \g w^\top x_i,\lambda^2)} - \E_q\sq{\log q(\eta_i)}.
\end{align}
The latent variables are $\eta_i$, the per-data point natural
parameters.  Their priors are Gaussians, each with mean $w^\top x_i$
and variance $\lambda^2$.

In an approximate E-step, we hold the parameters $w$ and $\lambda^2$
fixed and approximate the per-data point posterior $p(\eta_i \g y_i,
x_i, w, \lambda^2)$.  In theory, the optimal variational
distribution~\citep{Bishop:2006} is
\begin{align*}
  q(\eta_i) \propto \exp\left ( \eta_i y_i - a(\eta_i) + \log p(\eta_i
  \g w^\top x_i, \alpha)\right).
\end{align*}
But this does not easily normalize.

We address the problem with Laplace variational
inference~\citep{Wang:2013}. Laplace variational inference
approximates the optimal variational distribution with
\begin{align}\label{eq:laplace-var-inf}
  q(\eta_i) \approx \mathcal{N}(\hat{\eta}_i, -h^{-1}(\hat{\eta}_i)).
\end{align}
The value $\hat{\eta}_i$ maximizes the following function,
\begin{align}
  f(\eta_i) = \eta_i y_i - a(\eta_i) + \log p(\eta_i \g w^\top x_i, \lambda^2),
\end{align}
where $h(\cdot)$ is the Hessian of that function. Finding the
$\hat{\eta}_i$ can be done using many off-the-shelf optimization
routines, such as conjugate gradient.

Given these approximations to the variational distribution, the M-step
estimates $w$ and $\alpha$,
\begin{align}\label{eq:w-alpha-update}
  [\hat{w}, \hat{\alpha}] = \arg \max_{w,\alpha} \sum_{i=1}^n\expect{q(\eta_i)}{\log p(\eta_i \g w^\top x_i, \alpha)}.
\end{align}
In robust GLMs, the prior is Gaussian and we can compute the
expectation in closed form.   In general nonconjugate models, however,
we may need to approximate the expectation. Here we use the
multivariate delta method to approximate the
objective~\citep{Bickel:2007,Wang:2013}. \Cref{alg:robust-glm-alg}
shows the algorithm. 

\begin{algorithm}[!tb]
  \caption{Variational EM for a robust  GLM. \label{alg:robust-glm-alg}}
  \SetAlgoLined
  \DontPrintSemicolon
  \BlankLine
  \BlankLine
  Initialize $w$ and $\alpha$.\\
  \Repeat{the ELBO converges.}
  {
    \For{$i \in \{1,2,\dots,n\}$}{
      Update variational distribution $q(\eta_i)$
      (\Cref{eq:laplace-var-inf}).
    }
    Update $w$ and $\alpha$ using gradient ascent (\Cref{eq:w-alpha-update}). 
  }
\end{algorithm}

\parhead{Complex models with local and global variables.}  We can also
use variational inference when we localize more complex models, such
as mixture models or topic models.  Here we outline a strategy that
roughly follows~\cite{Hoffman:2013}.

We discussed complex Bayesian models in \Cref{sec:complex-models}; see
\Cref{fig:robust-latent-gm}.  Observations are $x_{1:n}$ and local
latent variables are $z_{1:n}$ and $\beta_{1:n}$.  (We have localized
the global variable $\beta$.)  The joint distribution is
\begin{align}
  p(\beta_{1:n}, x_{1:n}, z_{1:n}, \g \alpha ) =\textstyle
  \prod_{i=1}^{n} p(\beta_i \g \alpha) p(z_i,
  x_i \g \beta_i).\label{eq:joint}
\end{align}
Assume these distributions are in the exponential family,
\begin{align}
  p(z_i, x_i\g\beta_i) & \textstyle = h_{\ell}(z_i, x_i) \exp\left \{
    \beta_i^\top t_{\ell}(z_i, x_i) - a_{\ell}(\beta_i)
  \right \} \label{eq:conjugate-0}\\
  p(\beta_i \g \alpha) & \textstyle = h(\beta_i) \exp\left \{
    \alpha^\top t(\beta_i) - a(\alpha) \right \} \label{eq:conjugate},
\end{align}
The term $t(\beta_i)$ has the form
$t(\beta_i)= [\beta_i, -a_{\ell}(\beta_i)]$.  It is conjugate to
$p(z_i, x_i \g \beta_i)$.

This model satisfies \textit{conditional conjugacy}. The conditional
posterior $p(\beta_i \g z_i, x_i)$ in the same family as the prior
$p(\beta_i \g \alpha)$.  We emphasize that this differs from classical
Bayesian conjugacy---when we marginalize out $z_i$ the posterior of
$\beta_i$ is no longer in the same family.

The goal is to find the optimal $\alpha$ that maximizes the ELBO,
\begin{align}
  \mathcal{L}(\alpha) &= \sum_{i=1}^n\expect{q}{\log
                        p(x_i, z_i\g \beta_i)} + \expect{q}{p(\beta_i \g \alpha)} - \expect{q}{\log
                        q(z_{1:n}, \beta_{1:n})}
                  \label{eq:complex-elbo},
\end{align}
where the distribution $q(z_{1:n}, \beta_{1:n})$ contains both types
of latent variables.

We specify $q(\cdot)$ to be the mean-field family. It assumes a
fully factorized distribution,
\begin{align}
  \label{eq:meanfield}
  q(z_{1:n}, \beta_{1:n}) &= \textstyle \prod_{i=1}^{n} q(\beta_i) q(z_i).
\end{align}

% XXX no need to say it, but note this is an approximation.  z_i and
% beta_i are not necessarily independent in the exact posterior [dmb]

In the E-step we optimize the variational distribution.  We iterate
between optimizing $q(z_i)$ and $q(\beta_i)$ for each data point.
Because of conditional conjugacy, these updates are in closed form,
\begin{align}
  q(\beta_i) &\propto h(\beta_i) \exp\left ((\alpha +
  [\expect{q(z_i)}{t(z_i,x_i)}, 1])^\top t(\beta_i)\right
  ),\label{eq:q-beta} \\
  q(z_i) &\propto h(z_i,x_i) \exp\left (([t(z_i,x_i), 1])^\top
  \expect{q(\beta_i)}{t(\beta_i)}\right ).\label{eq:q-z}
\end{align}
Each $q(\cdot)$ will be in the same exponential family as its complete
conditional.  For fixed $\alpha$, the variational distribution
converges as we iterate between these updates.

In the M-step, we plug the fitted variational distributions
into~\Cref{eq:complex-elbo} and optimize $\alpha$.
\Cref{alg:robust-latent-alg} shows the algorithm.  This general method
fits robust versions of complex models, such as bursty
topic models or robust mixture models. 

\begin{algorithm}[!tb]
  \caption{Variational EM for robust models with local and global
    variables. \label{alg:robust-latent-alg}}
  \SetAlgoLined
  \DontPrintSemicolon
  \BlankLine
  \BlankLine
  Initialize $\alpha$.\\
  \Repeat{the ELBO converges.}
  {
    \For{$i \in \{1,2,\dots,n\}$}
    {
      Update $q(\beta_i)$ (\Cref{eq:q-beta}).\\
      Update $q(z_i)$ (\Cref{eq:q-z}).\\
    }
    Plug $q(\beta_i)$ and $q(z_i)$  into \Cref{eq:complex-elbo} and
    update parameter $\alpha$ with gradient ascent.
  }
\end{algorithm}

\section{Empirical Study}
\label{sec:study}

We study two types of robust Bayesian models---robust generalized
linear models and robust topic models. We present results on both
simulated and real-world data.  We use the strategy of
\Cref{sec:algorithms} for all models.  We find robust models
outperform their non-robust counterparts.

\subsection{Robust generalized linear models}

We first study the robust generalized linear models (GLMs) of
\Cref{sec:glm} and \Cref{sec:overdispersed}---linear regression,
logistic regression, and Poisson regression.  Each involves modeling a
response variable $y_i$ conditional on a set of covariates $x_i$.  The
response is governed (possibly through a localized variable) by a
linear combination with coefficients $w^\top x_i$.

We study robust GLMs with simulated data.  Our goal is to determine
whether our method for robust modeling gives better models when the
training data is corrupted by noise.  The idea is to fit various
models to corrupted training data and then evaluate those models on
uncorrupted test data.

Each simulation involves a problem with five covariates.  We first
generated true coefficients $w$ (a vector with five components) from a
standard normal; we then generated 500 test data points $(y_i, x_i)$
from the true model.  For each data point, the five covariates are
each drawn from $\mathrm{Unif}[-5,5]$ and the form of the response
depends on which model we are studying.  Next, we generate corrupted
training sets, varying the amount of corruption.  (How we corrupt the
training set changes from problem to problem; see below.)  Finally, we
fit robust and non-robust models to each training set and evaluate
their corresponding predictions on the test set. We repeat the
simulation 50 times.

We found that robust GLMs form better predictions than traditional
GLMs in the face of corrupted training data.\footnote{We compared our
  methods to the R implementations of traditional generalized linear
  models.  Linear, logistic, and Poisson regression are implemented in
the GLM package; negative binomial regression is in the MASS
package~\citep{Venables:2002}.}  Further, as expected, the performance
gap increases as the training data is more corrupted.

\parhead{Linear regression.}  We first use simulated data to study
linear regression.  In the true model
\[ y_i \g x_i \sim \mcN(w^\top x_i + b, 0.02). \]  
In the corrupted training data, we set a noise level $k$ and generate
data from \[ y_i \sim \mcN(w^\top x_i + b, \sigma_i + 0.02),\] where
$\sigma_i \sim \mathrm{Gamma}(k, 1)$.  As $k$ gets larger, there are
more outliers.  We simulated training sets with different levels of
outliers; we emphasize the test data does not include outliers.

\begin{figure}[ht]
\begin{center}
\centerline{\includegraphics[width=1.2\textwidth]{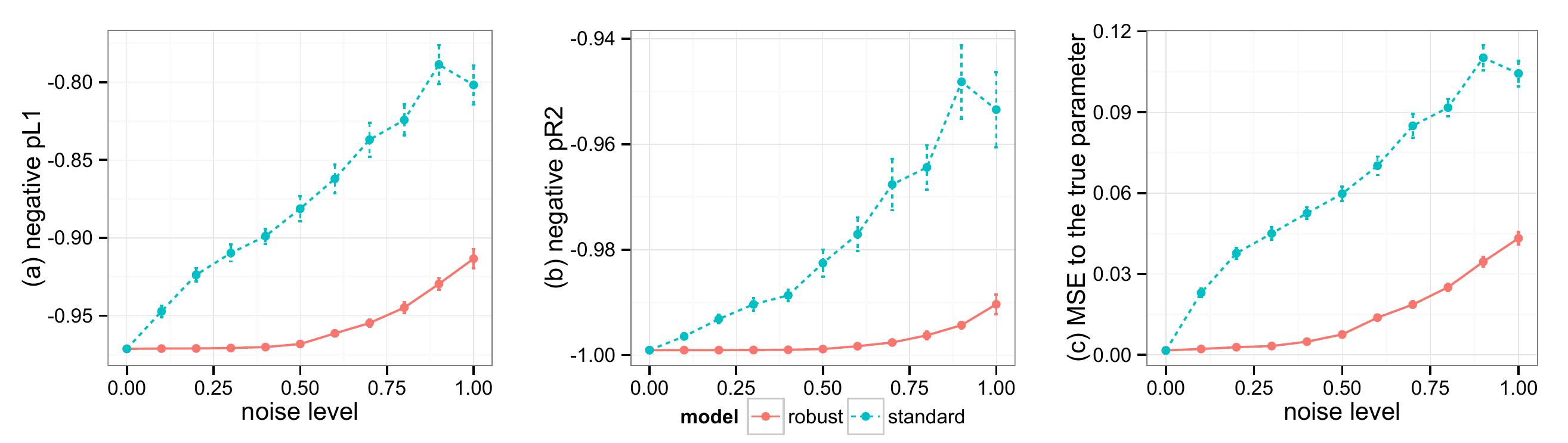}}
\caption{Robust linear regression compared to classical regression on
  simulated data.  The x-axis is additional noise in the training
  data, not captured by the model or present in test data.  Robust
  models perform better in the face of outliers. (a) Negative
  predictive L1; (b) Negative predictive R2. (c) MSE to the true
  parameter. For all metrics, lower is better.}

\label{fig:linear}
\end{center}
\end{figure}

We compare robust linear regression (with our general algorithm) to
standard regression.  After fitting coefficients $\hat{w}$ under the
robust model, we form predictions on test data as for linear
regression $\hat{y}_\textrm{new} = \hat{w}^\top x_\mathrm{new}$.  We
evaluate performance using three metrics: predictive L1,
\[\textstyle \mathrm{pL1} \triangleq 1-(\sum |y-\hat{y}|)/(\sum |y|),\]
predictive R2, \[\textstyle \mathrm{pR2} \triangleq 1-(\sum (y-\hat{y})^2)/(\sum
y^2),\] and the mean squared error to the true parameter (MSE) \[
  \textstyle \mathrm{MSE} \triangleq (1/d)\sum_{i=1}^d(\hat{w}_i-w_i)^2,\] where
$d$ is the dimension of parameter $w$. Figure~\ref{fig:linear} shows
the results.  The robust model is better than standard linear
regression when the training data is corrupted.  This is consistent
with the findings of~\citet{Lange:1989} and \citet{Gelman:2003}.

% QQQ in the original text you say that you set the class equal to
% I(w\top x_i + b > 0).  does that mean we didn't draw from the
% corresponding bernoulli?  (i think we should.)
% FIXQQQ: agreed.

\parhead{Logistic regression.} We next study logistic regression.  In
the true model, \[y_i \g x_i \sim \mathrm{Bernoulli}(\sigma(w^\top
x_i)),\] where $\sigma(\cdot)$ is the logistic function.  To contaminate
the training data, we randomly flip a percentage of the true labels,
starting with those points close to the true decision boundary.

\begin{figure}[t]
\begin{center}
\centerline{\includegraphics[width=1.2\textwidth]{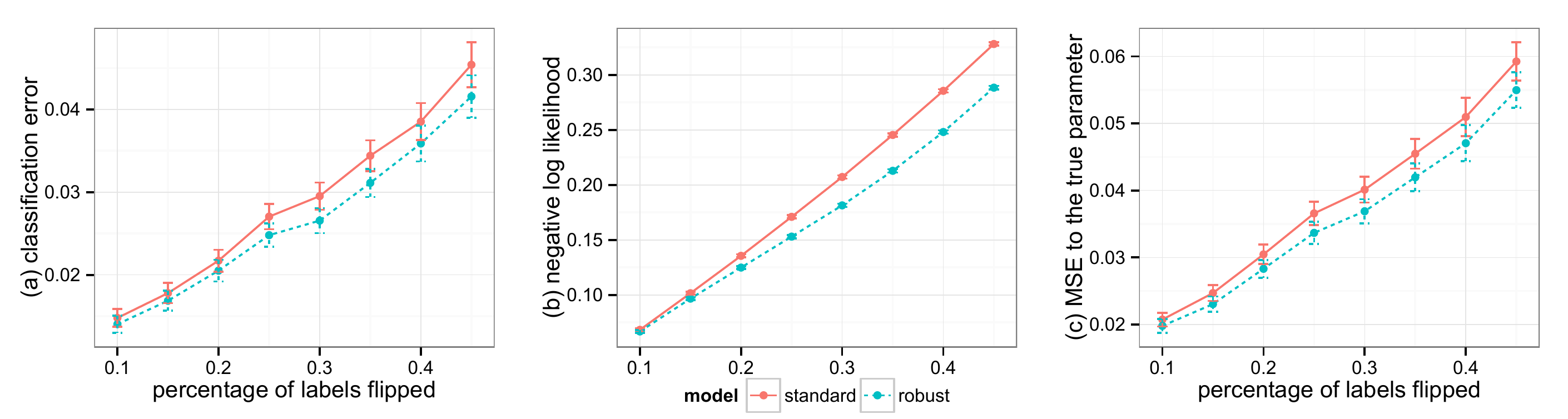}}
\caption{Experimental results for (robust) logistic regression on
    simulated data. Noise level (X-axis) indicates the proportion of
    the labels in the training data are flipped.  (a) Classification
    error; (b) Negative predictive log likelihood. (c) MSE to the true
parameter. All metrics: the {\it lower} the better. Robust model
perform better when noise is presented.}
\label{fig:logistic}
\end{center}
\end{figure}

We compare robust logistic regression
of~\Cref{eq:robust-poisson-1,eq:robust-poisson-2} to traditional
logistic regression.  Figure~\ref{fig:logistic} shows the results.
Robust models are better than standard models in terms of three
metrics: classification error, negative predictive log likelihood, and
mean square error (MSE) to the true data generating parameter $w$.

% QQQ i'm confused.  the original text suggests there is an epislon in
% the true data.  rather y_i ~ poisson(exp(w^\top x_i). [dmb]
% FIXQQQ: i don't think there is an epislon in the true data, see page
% 9?

\parhead{Poisson regression.} Finally we study Poisson regression.  In
the true model
\[y_i \g x_i \sim \mathrm{Poisson}\left(\exp\left\{w^\top
x_i\right\}\right).\]
We corrupt the training data by sampling a per-data point noise
component $\epsilon_i \sim \mcN(0, \sigma^2)$ and then generating data
from
\[y_i \g x_i \sim \mathrm{Poisson}\left(\exp\left\{w^\top x_i +
\epsilon_i\right\}\right).\]
The variance $\sigma^2$ controls the amount of noise in the training
data.

\begin{figure}[t]
\begin{center}
\centerline{\includegraphics[width=1.2\textwidth]{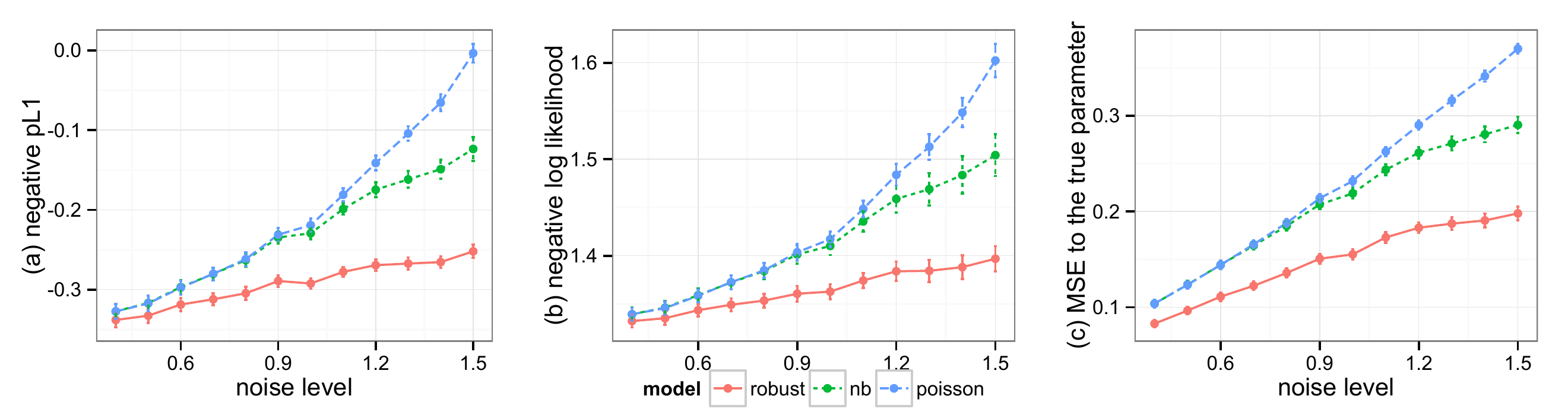}}
\caption{Experimental results for (robust) Poisson regression on
    simulated data. (a) Negative predictive L1; (b) Negative
    predictive log likelihood. (c) MSE to the true parameter. All
    metrics: the {\it lower} the better. Robust models tend to perform
better than both Poisson regression and negative binomial regression
(nb) when noise is presented.} 

\label{fig:poisson-pL1}
\end{center}
\end{figure}

We compare our robust Poisson regression to traditional Poisson
regression and to negative binomial regression.
Figure~\ref{fig:poisson-pL1} shows the results.  We used three
metrics: predictive L1 (as for linear regression), negative predictive
log likelihood, and MSE to the true coefficients.  Robust
models are better than both standard Poisson regression and the
negative binomial regression, especially when there is large noise.

Note that negative binomial regression is also a robust model.  In a
separate study, we confirmed that it is the empirical Bayes step,
where we fit the variance around the per-data point parameter, that
explains our better performance.  Using the robust Poisson model
without fitting that variance (but still fitting the coefficients)
gave similar performance to negative binomial regression.

\begin{figure}[t]
\begin{center}
    \centerline{\includegraphics[width=1.2\textwidth]{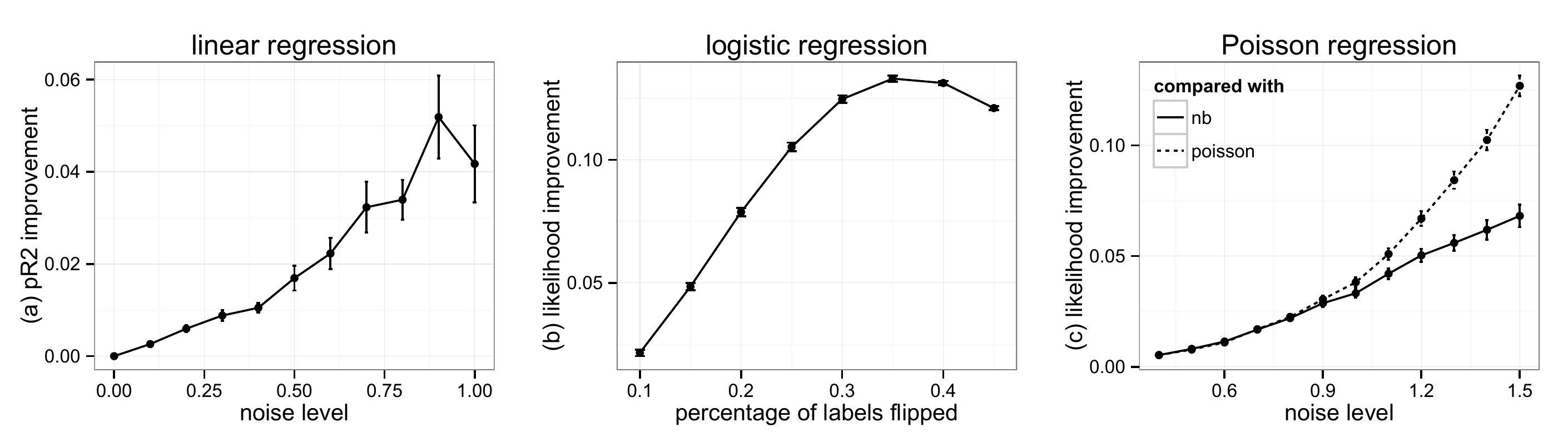}}
\caption{Improvement of robust models over standard models. (a)
Predictive R2 improvement for linear regression; (b) log likelihood
improvement for logistic regression (c) log likelihood improvement for
Poisson regression. } 

\label{fig:improvement}
\end{center}
\end{figure}

\parhead{Summary.} We summarize these experiments with
Figure~\ref{fig:improvement} showing the improvement of robust models
over standard models in terms of log likelihood. (For linear
regression, we use pR2.) Robust models give greater improvement when
the data is noisier.

\subsection{Robust topic modeling}
We also study robust LDA, an example of a complex Bayesian model
(\Cref{sec:complex-models}).  We have discussed that robust LDA is a
bursty topic model~\citep{Doyle:2009}.

We analyze three document corpora: {\it Proceedings of the National
  Academy of Sciences} (PNAS), {\it Science}, and a subset of {\it
  Wikipedia}. The PNAS corpus contains 13,000 documents and has a
vocabulary of 7,200 terms; the {\it Science} corpus contains 16,000
documents and has a vocabulary of 4,400 terms; the {\it Wikipedia}
corpus contains about 10,000 documents and has a vocabulary
of 15,300 terms.  We run a similar study to the one
in~\cite{Doyle:2009}, comparing robust topic models to traditional
topic models.

\parhead{Evaluation metric.}  To evaluate the methods, we hold out
20\% documents from each corpus and calculate their predictive
likelihood.  We follow the metric used in recent topic modeling
literature~\citep{Blei:2007,Asuncion:2009,Wang:2011a,Hoffman:2013},
where we hold out part of a document and predict its remainder.

Specifically, for each document in the test set $\bw_d$, we split it
in into two parts, $\bw_d = [\bw_{d1}, \bw_{d2}]$. We compute the
predictive likelihood of $\bw_{d2}$ given $\bw_{d1}$ and
$\mathcal{D}_{\rm train}$.  The per-word predictive log likelihood is
\begin{align*}
  {{\rm likelihood}_{\rm pw}} \triangleq  \frac{\sum_{d \in
      \mathcal{D}_{\rm test}}\log p( \bw_{d2}| \bw_{d1},
    \mathcal{D}_{\rm train})}{ \sum_{ d \in \mathcal{D}_{\rm test}}
    | \bw_{d2}| },
\end{align*}
where $|\bw_{d2}|$ is the number of tokens in $\bw_{d2}$.  This
evaluation measures the quality of the estimated predictive
distribution. This is similar to the strategy used in~\citet{Hoffman:2013}.

For standard LDA~\citep{Blei:2003b}, conditioning on $\bw_{d1}$
estimates the topic proportions $\theta_d$ from corpus-wide topics.
These topic proportions are then used to compute the predictive
likelihood of $\bw_{d2}$. Robust LDA is different because conditioning
on $\bw_{d1}$ estimates both topic proportions and per-document
topics; the predictive likelihood of $\bw_{d2}$ uses both quantities.

% We compare robust topic models to traditional topic models.  In robust
% topic models we use the per-document topics when finding the
% predictive distribution conditional on $\bw_{d1}$; in traditional
% topic models we use the fitted topics.

\begin{figure}[t]
\begin{center}
\centerline{\includegraphics[width=1.2\textwidth]{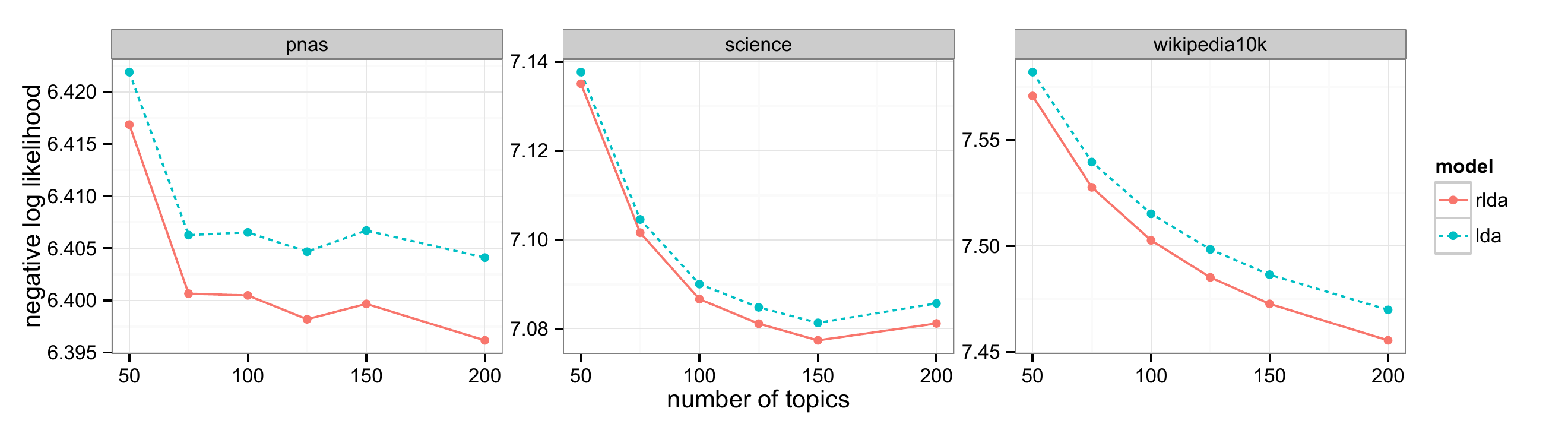}}
\caption{Predictive negative log likelihood on test data. The lower
the better. The robust model (rlda) also performs better over a range
of topics.} \label{fig:lda-rlda}
\end{center}
\end{figure}

\parhead{Results.} \Cref{fig:lda-rlda} shows the results. (Note, in
the figure we use negative log likelihood so that it is consistent
with other plots in this paper.)  Robust topic models perform better
than traditional topic models. This result is consistent with those
reported in~\cite{Doyle:2009}.

\section{Summary}
We developed a general method for robust Bayesian modeling.
Investigators can create a robust model from a standard Bayesian model
by localizing the global variables and then fitting the resulting
hyperparameters with empirical Bayes; we described a variational EM
algorithm for fitting the model. We demonstrated our approach on
generalized linear models and topic models.

\bibliographystyle{ba}
\bibliography{bib}

\begin{acknowledgement}
David M. Blei is supported by NSF BIGDATA NSF IIS-1247664, NSF NEURO
  NSF IIS-1009542, ONR N00014-11-1-0651, DARPA FA8750-14-2-0009,
  Facebook, Adobe, Amazon, the Sloan Foundation, and the John
  Templeton Foundation.  The authors thank Rajesh Ranganath and Yixin
  Wang for helpful discussions about this work.
\end{acknowledgement}

\end{document}